\documentclass{article}

    \PassOptionsToPackage{numbers, compress}{natbib}
\usepackage[preprint]{neurips_2025}

\usepackage[utf8]{inputenc} 
\usepackage[T1]{fontenc}    
\usepackage{hyperref}       
\usepackage{url}            
\usepackage{booktabs}       
\usepackage{amsfonts}       
\usepackage{nicefrac}       
\usepackage{microtype}      
\usepackage{xcolor}         
\usepackage{multirow}

\usepackage{graphicx}
\usepackage{amsmath}
\usepackage{adjustbox}
\usepackage{xspace}
\newcommand{\model}{ZR-0\xspace}

\hypersetup{hidelinks}

\title{Training Vision-Language-Action Models with Dense Embodied Chain-of-Thought Supervision}

\author{%
  Haoyang Li\textsuperscript{1,}\thanks{Work partially done during an internship at Zhipu AI.} \quad
  Guanlin Li\textsuperscript{1,*} \quad
  Youhe Feng\textsuperscript{1,*} \quad
  Chen Zhao\textsuperscript{1,*} \quad
  Zhuoran Wang\textsuperscript{1,*} \\
  \textbf{Yang Li\textsuperscript{1,*} \quad
  Qizhe Wei\textsuperscript{1} \quad
  Shifeng Bao\textsuperscript{1,*}} \\
  \textbf{Haitao Shen\textsuperscript{1} \quad
  Yihan Zhao\textsuperscript{1} \quad
  Tong Yang\textsuperscript{2} \quad
  Jing Zhang\textsuperscript{1}\thanks{Corresponding author.}} \\[0.5em]
  \textsuperscript{1}Renmin University of China \quad
  \textsuperscript{2}Zhipu AI
}

\begin{document}

\maketitle

\begin{abstract}
Cross-embodiment transfer in vision-language-action (VLA) models remains challenging because low-level state and action spaces differ fundamentally across robot platforms. We observe that the high-level cognitive process underlying manipulation, including scene perception, object identification, task planning, and sub-task decomposition, is largely shared across embodiments. Based on this observation, we present \textbf{\model}, a 2.6 billion parameter end-to-end VLA model that uses dense Embodied Chain-of-Thought (ECoT) supervision to align cross-embodiment representations within the vision-language model (VLM). \model adopts a dual-stream architecture: a pre-trained VLM (System 2) generates structured ECoT reasoning during training, while a Diffusion Transformer-based action expert (System 1) produces continuous action chunks via flow matching. The two components are coupled through cross-attention, with an attention mask that restricts the action expert to input prompt features only, enabling ECoT generation to be entirely skipped at inference without any performance loss. \model is pre-trained on ProcCorpus-60M, a large-scale dataset comprising approximately 60 million frames (approximately 1,000 hours) from over 400K trajectories, with dense ECoT annotations covering 96.8\% of all frames. We evaluate \model on three simulation benchmarks spanning single-arm (LIBERO), bimanual (RoboTwin 2.0), and humanoid (RoboCasa GR-1 Tabletop) embodiments, as well as real-world experiments on the xArm platform, demonstrating strong performance across all settings. Code and model checkpoints are available at \url{https://github.com/RUCKBReasoning/ZR-0}.
\end{abstract}

\section{Introduction}

Building generalist robots capable of performing diverse manipulation tasks across different embodiments is a central goal of embodied AI. Inspired by the success of large-scale pre-training in natural language processing and computer vision, the robotics community has increasingly adopted vision-language-action (VLA) models~\citep{black2024@pi0, bjorck2025@GR00T, Cheang2025@GR-3, google2025@GeminiRobotics, kim2024@openvla} as a paradigm for learning general-purpose robotic policies. By pre-training on large-scale robotics datasets aggregated from diverse sources~\citep{oneill2024@oxe, Khazatsky2024@droid, bu2025@AgiBotWorld, fang2024@RH20T}, these models aim to acquire transferable physical commonsense and manipulation skills that can be efficiently adapted to new tasks, scenes, and robot embodiments.

A key promise of this paradigm lies in \emph{cross-embodiment transfer}: training a single model on data from many heterogeneous robots, so that knowledge learned from one embodiment benefits others. However, achieving effective cross-embodiment transfer remains a fundamental challenge. Different robot platforms vary substantially in their kinematic configurations (e.g., 6-DoF vs.\ 7-DoF arms), control interfaces (e.g., joint position vs.\ end-effector pose with varied rotation representations), base types (e.g., fixed-base vs.\ mobile), and sensor setups. These differences manifest as heterogeneous state and action spaces, where individual dimensions carry different physical meanings across embodiments. Existing approaches address this primarily through format-level techniques such as zero-padding and per-embodiment normalization~\citep{black2024@pi0, bjorck2025@GR00T}. Other methods attempt to define a unified action space by assigning fixed semantic roles to each dimension~\citep{Liu2025@rdt, Apanasevich2026@greenvla, yuan2026@qwenrobomanip}. However, even when actions are placed into corresponding dimensions, the same dimension (e.g., joint 1) can carry different physical meanings across embodiments, since the rotation axis and range of each joint differ from one robot to another. These format-level solutions enable joint training but do not resolve the deeper challenge of \emph{semantic alignment}: ensuring that the model learns shared, transferable representations rather than merely fitting embodiment-specific patterns within a unified architecture.

While low-level state and action spaces are inherently embodiment-specific, the high-level cognitive process underlying manipulation, such as perceiving the scene, reasoning about task progress, planning the next steps, and identifying target objects, is largely \emph{shared} across embodiments. A robot arm picking up a cup from a table follows a similar cognitive trajectory regardless of whether the arm has 6 or 7 degrees of freedom. This embodiment-agnostic reasoning constitutes the transferable knowledge that cross-embodiment pre-training should capture.

Based on this observation, we present \textbf{\model}, a 2.6 billion parameter end-to-end VLA model that leverages \textbf{Embodied Chain-of-Thought (ECoT)} reasoning as a dense supervision signal to align cross-embodiment representations. \model adopts a dual-stream architecture inspired by the System 1/System 2 cognitive framework: \textbf{System 2}, a pre-trained vision-language model (VLM), processes visual observations and task instructions to produce structured ECoT reasoning that captures embodiment-agnostic understanding of the current scene and task; \textbf{System 1}, a Diffusion Transformer (DiT)-based action expert, takes the VLM representations and maps them to embodiment-specific continuous action chunks via flow matching. The two systems are connected through cross-attention, enabling rich information flow from reasoning to action.

Crucially, while ECoT supervision is used during training to drive the VLM to learn semantically aligned, transferable representations, ECoT text generation is \emph{entirely omitted at inference}. By applying a cross-attention mask that restricts the action expert to attend only to the VLM's input prompt features, a single forward pass of the VLM suffices to produce all features required by the action expert. This design retains the representational benefits of ECoT without incurring its inference cost.

\model is pre-trained on ProcCorpus-60M~\cite{Feng2026@procvlm}, a large-scale ECoT-enriched robotic dataset comprising approximately 60 million frames (approximately 1,000 hours) from over 400K trajectories across diverse embodiments. Each frame is annotated with a structured ECoT sequence consisting of a scene description, task progress assessment, future plan, decomposed atomic sub-task actions, target object bounding boxes, and discretized action tokens, collectively bridging high-level language instructions and low-level control in an embodiment-agnostic format. This dense ECoT supervision across heterogeneous embodiments is what enables \model to learn aligned, transferable representations.

We evaluate \model on three simulation benchmarks covering single-arm (LIBERO), bimanual (RoboTwin 2.0), and humanoid (RoboCasa GR-1 Tabletop) embodiments, as well as real-world experiments on the xArm platform. Results demonstrate that \model achieves strong performance across all settings, validating the effectiveness of dense ECoT supervision for cross-embodiment VLA training.

\section{Related Work}

\textbf{Vision-Language-Action (VLA) Models.}
Leveraging the rich visual and linguistic knowledge encoded in pretrained vision-language models (VLMs), vision-language-action (VLA) models~\cite{bjorck2025@GR00T, black2024@pi0, black2025@pi0.5, Cheang2025@GR-3, google2025@GeminiRobotics1.5, google2025@GeminiRobotics, Qu2025@EO1, Lee2025@MolmoAct, Fang2026@MolmoAct2, Zheng2025@xvla, pi2026@pi0.7, Yang2026@abot-m0, Cai2026@internvla-a1, yuan2026@qwenrobomanip, joy2026@joyai-ra,  Pertsch2025@pifast, Lin2025@OneTwoVLA, Zhai2025@wall-oss} have become a prominent paradigm for learning generalist robotic policies. Early VLA approaches, including RT-2~\cite{Zitkovich2023@RT-2}, OpenVLA~\cite{kim2024@openvla}, and FAST~\cite{Pertsch2025@pifast}, represent continuous actions as discrete tokens, thereby aligning action prediction with the autoregressive generation framework of VLMs. While this design enables straightforward integration with standard VLM training pipelines, it introduces sequential decoding overhead and can suffer from precision loss due to action tokenization and detokenization.

To address these limitations, $\pi_0$~\cite{black2024@pi0} proposes a Mixture-of-Transformers architecture that combines a pretrained VLM with a flow-matching-based action expert, allowing continuous action chunks to be predicted directly for high-frequency control. $\pi_{0.5}$~\cite{black2025@pi0.5} further augments this architecture with high-level subtask planning, improving long-horizon task execution and generalization. GR00T N1~\cite{bjorck2025@GR00T} continues this line of work by replacing the Mixture-of-Transformers-based action expert with a cross-attention-based Diffusion Transformer (DiT)~\cite{peebles2023@dit}, enabling more flexible combinations of VLM backbones and action experts. \model is also built on a dual-stream VLA architecture, but differs from these approaches by introducing dense ECoT supervision into the VLM stream to improve cross-embodiment representation learning.

\textbf{Vision-Language Data Co-training for Robot Learning.}
Robot demonstration data is costly to collect and limited in scale. In addition, fine-tuning on robot trajectories with action-only supervision can erode the general visual and linguistic capabilities inherited from pretrained VLMs, reducing policy generalization. To mitigate these issues, recent works co-train robotic data with auxiliary vision-language (VL) data~\cite{Lee2025@MolmoAct, driess2025@KI, black2025@pi0.5, yuan2026@qwenrobomanip, joy2026@joyai-ra, Qu2025@EO1, Lin2026@vl-cotrain, zhou2025@chatvla-2, Yang2025@Vlaser, Lin2025@OneTwoVLA, Zhai2025@wall-oss, chen2025@efficient_ecot} to preserve broad VLM knowledge while adapting models to robotic control.

Some approaches use general-purpose VL corpora, including visual question answering, image captioning, OCR, and visual grounding datasets~\cite{Deitke2025@molmo-and-pixmo, Lin2014@coco, Tong2024@Cambrian, Yu2024@capsfusion, Yuan2024@RoboPoint, Liu2023@LLAVA}, to maintain broad visual and language understanding during VLA fine-tuning. Others construct \emph{embodied reasoning} VL data directly from robot trajectories~\cite{zawalski2024@ecot, Lin2025@OneTwoVLA, Zhai2025@wall-oss, yuan2026@qwenrobomanip, ji2025robobrain, Yang2025@Vlaser}, providing fine-grained supervision such as scene descriptions, spatial understanding, subtask prediction, and motion planning. Because such supervision is tightly coupled to the manipulation context, it is often more directly useful for downstream action prediction. The ECoT framework used in \model follows this second line of work, but focuses on aligning VLM representations across heterogeneous embodiments through dense reasoning supervision at scale.

\section{\model}

We present \model, a 2.6 billion parameter end-to-end Vision-Language-Action (VLA) model that combines embodied chain-of-thought (ECoT) reasoning with diffusion-based action generation. \model is designed to accommodate diverse robot embodiments, ranging from single arms (e.g., Franka, XArm) to bimanual platforms (e.g., AgiBot G1, Agilex). To this end, \model is trained on ProcCorpus-60M~\cite{Feng2026@procvlm}, a large-scale ECoT-enhanced robotic dataset constructed by aggregating several major open-source robot datasets (including Open X-Embodiment~\citep{oneill2024@oxe}, DROID~\citep{Khazatsky2024@droid}, RH20T~\citep{fang2024@RH20T}, and more) and automatically annotating every frame with structured ECoT reasoning via a dedicated pipeline. ProcCorpus-60M spans a diverse range of tasks, scenes, embodiments, and behaviors, enabling \model to acquire generalizable and transferable physical commonsense for robot control.

\subsection{Model Architecture}\label{sec:architecture}

\begin{figure*}[htbp]
    \centering
    \includegraphics[width=0.99\textwidth]{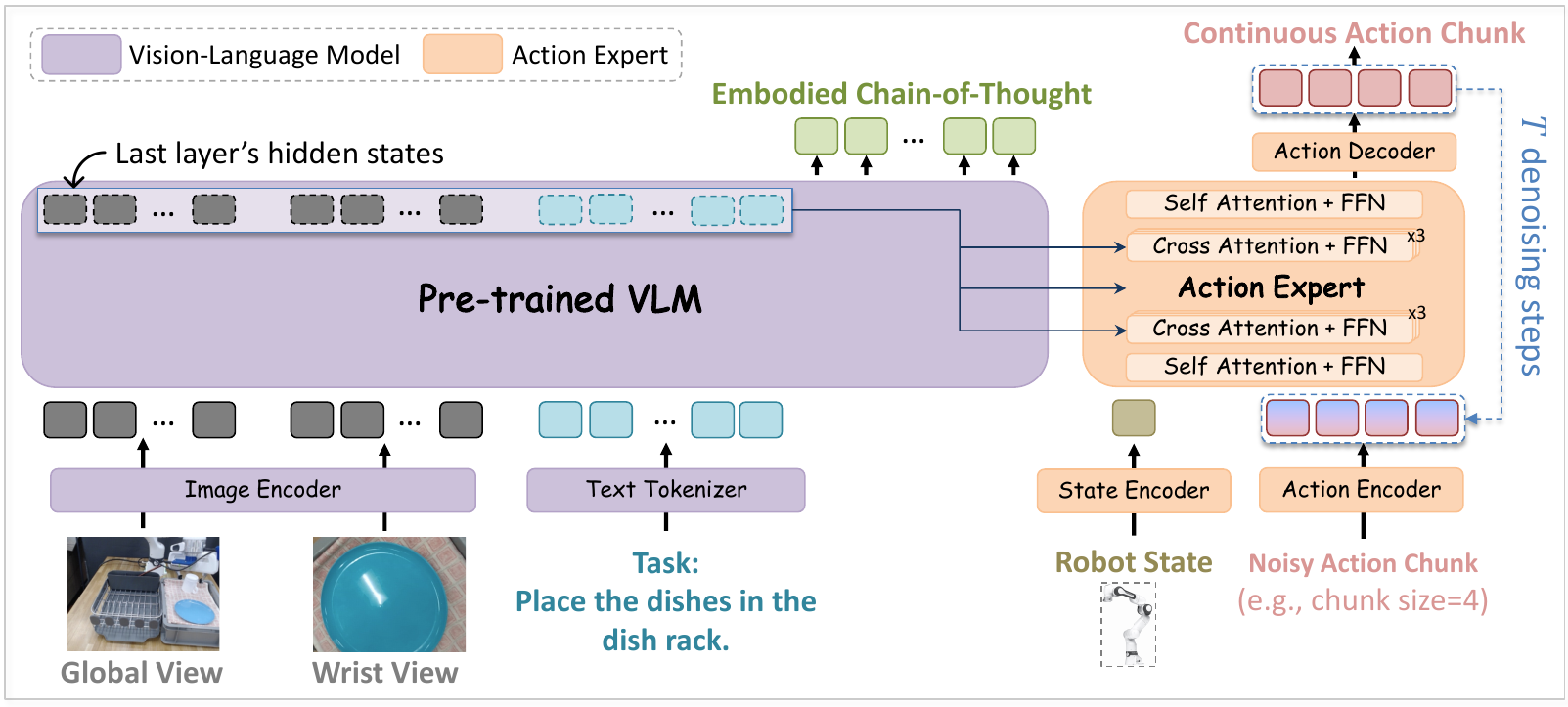}
    \caption{The framework of \model. \model combines a vision-language model (VLM) with a Diffusion Transformer (DiT)-based action expert. Joint training is performed on embodied chain-of-thought (ECoT) via next-token prediction, and on continuous actions using denoising vector field prediction.}
    \label{fig:model_architecture}
\end{figure*}

As illustrated in Figure~\ref{fig:model_architecture}, \model adopts a dual-stream architecture inspired by the System~1 / System~2 cognitive framework. \textbf{System~2} is a pre-trained vision-language model (VLM) that processes task instructions and image observations to produce structured ECoT reasoning. \textbf{System~1} is a Diffusion Transformer (DiT)-based action expert that generates a chunk of $H$ continuous actions via flow matching. The two components are coupled through cross-attention, allowing the action expert to condition on the VLM's representations. This architecture flexibly integrates high-level reasoning with low-level continuous control, enabling both interpretable decision-making and precise, high-frequency action generation.

\textbf{Vision-Language Model (System~2).}
Pre-trained on web-scale multimodal data, VLMs encode rich visual and linguistic knowledge that provides a strong foundation for robotic policy learning~\citep{kim2024@openvla, black2024@pi0, bjorck2025@GR00T, black2025@pi0.5, Cheang2025@GR-3}. In \model, the VLM is initialized from Qwen3-VL-2B-Instruct~\citep{bai2025@Qwen3-VL}. Given a natural language task instruction $l$ and image observations $o_{t} = [img_{t}^{1}, \ldots, img_{t}^{n}]$ from $n$ camera views at timestep $t$, the VLM is trained to generate an ECoT reasoning sequence $r_{t}$.

Prior work~\citep{chen2025@efficient_ecot, zawalski2024@ecot} has shown that ECoT supervision provides rich gradient signals that improve the VLM's learned representations, which in turn benefit downstream action prediction. In practice, we extract features from the last-layer hidden states of the VLM, denoted $f_{t}$, and pass them to the action expert. All input images are resized to $224 \times 224$.

\textbf{Diffusion Transformer-based Action Expert (System~1).}
To model actions in continuous spaces, we employ a variant of the Diffusion Transformer (DiT)~\citep{peebles2023@dit} as the action expert. Given VLM features $f_{t}$ and the robot state vector $s_t$, the action expert is trained via flow matching to predict an action chunk $A_{t} = [a_{t}, a_{t+1}, \ldots, a_{t+H-1}]$. As shown in Figure~\ref{fig:model_architecture}, the action expert consists of a state encoder, an action encoder, a stack of DiT blocks, and an action decoder, where the encoders and decoder are implemented as MLPs.

For VLM feature integration, the DiT blocks follow a repeating pattern of one self-attention layer followed by three cross-attention layers. In self-attention layers, bidirectional attention is applied between state and action tokens to facilitate feature fusion. In cross-attention layers, state and action tokens serve as queries while the VLM's output features serve as keys and values. Crucially, we apply an attention mask that restricts the action expert to attend only to the VLM's features corresponding to the input prompt (i.e., task instruction and images), excluding the ECoT tokens. This design choice is what enables \model to skip ECoT generation entirely at inference: a single forward pass of the VLM over the input prompt suffices to produce all features required by the action expert, without the need for autoregressive ECoT decoding. Unlike the 1:1 self-attention-to-cross-attention ratio used in GR00T N1~\citep{bjorck2025@GR00T}, our 1:3 ratio increases the proportion of cross-modal interaction, allowing the action expert to more thoroughly absorb task instructions and visual observations from the VLM.

\subsection{Pre-Training Data}
\label{sec:training_data}

\textbf{ProcCorpus-60M.}
Since \model relies on ECoT supervision to learn cross-embodiment aligned representations, the training corpus must provide dense ECoT annotations across diverse embodiments. We adopt ProcCorpus-60M~\cite{Feng2026@procvlm} as the primary training corpus for \model. ProcCorpus-60M aggregates over 60 million frames (approximately 1,000 hours) from more than 400K trajectories sourced from a diverse collection of real-robot and simulated datasets, including DROID~\citep{Khazatsky2024@droid}, Bridge~\cite{Walke2023@Bridge}, Fractal~\cite{Brohan2023@RT-1}, RH20T~\citep{fang2024@RH20T}, several Open X-Embodiment subsets~\citep{oneill2024@oxe}, and others.
Critically, ProcCorpus-60M provides dense Embodied Chain-of-Thought (ECoT) annotations for nearly every frame (96.8\% annotation coverage), generated through an automated VLM-based annotation pipeline~\cite{Feng2026@procvlm}. This dense supervision across heterogeneous embodiments is what enables \model to learn aligned, transferable representations through ECoT.

\textbf{Components of ECoT and Their Roles.}
Each ECoT annotation is a structured sequence comprising six components, each designed to strengthen a specific aspect of the VLM's capabilities:

\begin{itemize}
    \item \textbf{Scene Description:} A textual depiction of the current visual scene. This component trains the VLM to improve object recognition capabilities, strengthening its ability to identify task-relevant objects in the workspace.

    \item \textbf{Progress Assessment:} A brief reasoning passage that evaluates what has been accomplished so far, followed by a binary completion indicator (Yes/No). This component trains the VLM to perceive task progress.

    \item \textbf{Future Plan:} A free-form natural language description reasoning about what remains to be accomplished to fulfill the instruction. This component trains the VLM to perform temporal reasoning and long-horizon planning based on the current observation and task progress.

    \item \textbf{To-Do Actions:} A structured decomposition of the future plan into a list of atomic sub-tasks, each expressed as an imperative sentence in the form \emph{Verb + Object [+ Prepositional Phrase]} (e.g., ``Grasp the blue plate from the towel.'', ``Place the blue plate into the dish rack.''). While the future plan captures the overall remaining intent in natural language, to-do actions refine it into fine-grained, executable steps. By expressing these sub-goals in an embodiment-agnostic format, this component serves as a key mechanism for cross-embodiment alignment, since the same sub-task decomposition applies regardless of the underlying robot hardware.

    \item \textbf{Target Objects:} Bounding boxes in standard JSON format localizing the object(s) relevant to the current manipulation step (e.g., \texttt{\{"blue plate": [120, 85, 340, 260]\}}). This visual grounding supervision directs the model's spatial attention toward task-critical regions, improving generalization across camera viewpoints and scene layouts.

    \item \textbf{Discrete Actions:} Embodiment-specific discrete action tokens produced by the FAST tokenizer~\citep{Pertsch2025@pifast}. These tokens provide a compact bridge between the high-level, embodiment-agnostic reasoning in the preceding ECoT components and the low-level continuous control of the action expert.
\end{itemize}

\textbf{Mixing General Vision-Language Data.}
In addition to robotic trajectory data, we mix general-purpose vision-language datasets, including CapsFusion~\citep{Yu2024@capsfusion} and Pixmo~\cite{Deitke2025@molmo-and-pixmo}, into the pre-training corpus.
These datasets cover tasks such as visual question answering, image captioning, and visual grounding. Unlike ECoT-annotated robot data, which provides supervision for both the VLM and the action expert, these pure VL data points are used to train the VLM only via standard language modeling, with no action prediction involved. This co-training strategy preserves the VLM's general visual perception and language understanding capabilities acquired during its original pre-training, mitigating catastrophic forgetting and thereby improving \model's robustness to novel scenes and its ability to follow diverse natural language instructions.

\subsection{Training Objective}

\model is jointly optimized with two complementary objectives: (1) next-token prediction for ECoT reasoning, and (2) denoising vector field prediction for continuous action generation.

For ECoT generation, we adopt a standard next-token prediction loss~\citep{Radford2018@gpt1}:
\[
\mathcal{L}_{\mathrm{ntp}} = - \mathbb{E}_{D} \left[ \sum_{i} \log \pi_{\theta'}(r_{t}^{i}\mid l, o_{t}, r_{t}^{<i}) \right],
\]
where $D$ denotes the training dataset, $\theta'$ the VLM parameters, $l$ the task instruction, $o_{t}$ the image observations, and $r_{t}^{i}$, $r_{t}^{<i}$ the $i$-th token and all preceding tokens in the ECoT sequence, respectively.

For continuous action chunk prediction, given a ground-truth action chunk $A_{t}$, Gaussian noise $\epsilon \sim \mathcal{N}(0, I)$, and a flow matching timestep $\tau \in [0, 1]$, we construct a noisy action chunk $A_{t}^{\tau} = (1-\tau)\epsilon + \tau A_{t}$. The whole model is trained to approximate the denoising vector field $A_{t} - \epsilon$ by minimizing:
\[
\mathcal{L}_{\mathrm{fm}} = \mathbb{E}_{D, \tau, \epsilon} \left[ \left\| \pi_\theta(l, o_{t}, s_{t}, A_{t}^{\tau}, \tau) - (A_{t} - \epsilon) \right\|^2 \right],
\]
where $l$ is the task instruction and $o_t$ the image observations. Following~\citet{black2024@pi0}, we sample $\tau$ from a $\mathrm{Beta}(1.5, 1.0)$ distribution to emphasize noisier timesteps during training. Internally, the VLM first encodes $l$ and $o_t$ into features $f_t$, and a cross-attention mask restricts the action expert to attend only to features corresponding to the input prompt (excluding ECoT tokens), as described in Section~\ref{sec:architecture}.

The overall loss is a weighted sum of the two objectives:
\[
\mathcal{L} = \mathcal{L}_{\mathrm{ntp}} + \alpha\,\mathcal{L}_{\mathrm{fm}},
\]
where $\alpha \in \mathbb{R}$ controls the trade-off. Notably, $\mathcal{L}_{\mathrm{ntp}}$ updates only the VLM parameters, while $\mathcal{L}_{\mathrm{fm}}$ propagates gradients through both the action expert and the VLM (via $f_t$).

\subsection{Inference}

At inference time, \model receives the task instruction $l$, image observations $o_t$, and robot state $s_t$ from the environment. A noisy action chunk is initialized from Gaussian noise, $A_{t}^{0} \sim \mathcal{N}(0, I)$, and iteratively refined via forward Euler integration:
\[
A_{t}^{\tau+\frac{1}{N}} = A_{t}^{\tau} + \frac{1}{N} \cdot \pi_\theta(l, o_{t}, s_{t}, A_{t}^{\tau}, \tau),
\]
where $N$ is the number of denoising steps and $\tau$ is initialized to $0$ and incremented by $1/N$ after each step. After $N$ iterations, $A_{t}^{1}$ yields the predicted action chunk.

Importantly, although ECoT is used as a training supervision signal, \model does not generate ECoT sequences at inference time. By bypassing costly autoregressive text generation, \model achieves substantially lower inference latency while retaining the representational benefits of ECoT~\citep{chen2025@efficient_ecot}. On a single NVIDIA A6000 GPU with \texttt{bfloat16} precision, generating an action chunk takes approximately 90\,ms, yielding an effective control frequency well suited for real-time deployment.

\section{Experiments}

We evaluate \model on three simulation benchmarks, LIBERO (single-arm), RoboTwin 2.0 (bimanual), and RoboCasa GR-1 Tabletop (humanoid), as well as real-world experiments on the xArm platform, covering diverse embodiments, tasks, and scene configurations.

\begin{figure*}[htbp]
    \centering
    \includegraphics[width=0.99\textwidth]{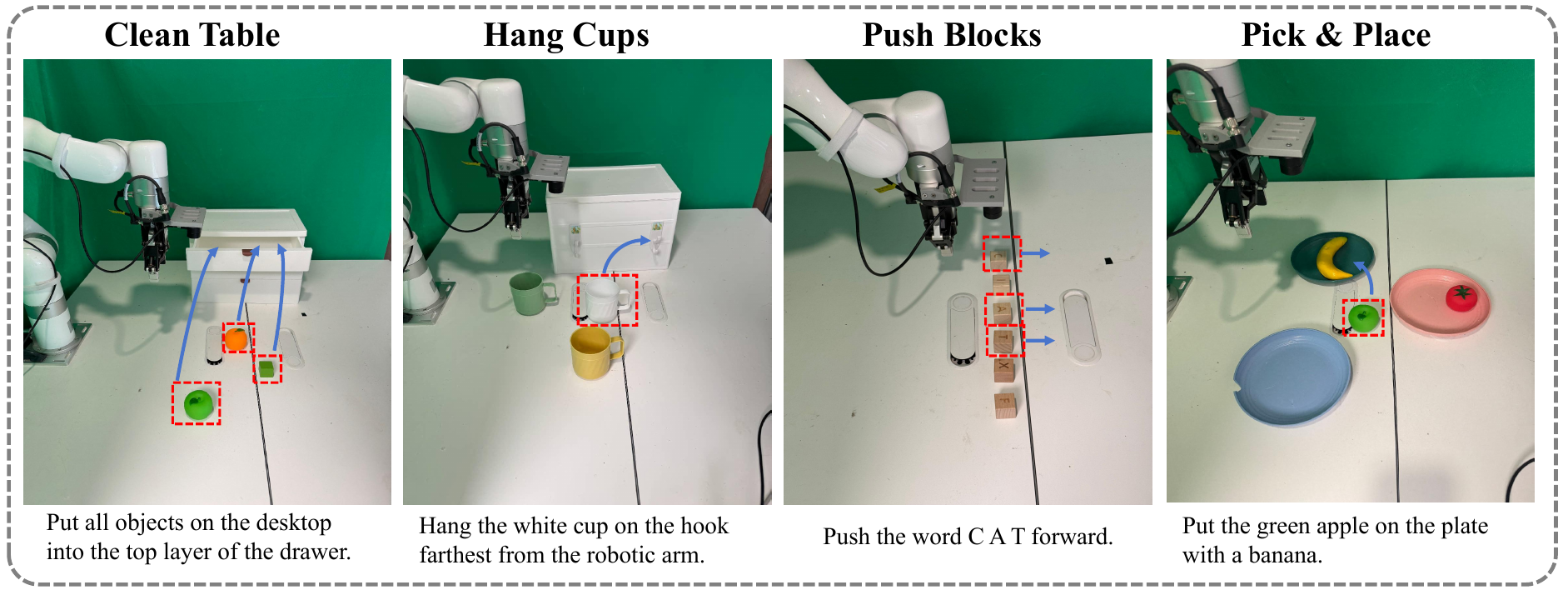}
    \caption{Examples of real-world robotic environments and task setups. We evaluate \model on four diverse tasks to assess its capabilities in instruction following, color understanding, long-horizon planning, spatial reasoning, and OCR-based reasoning.}
    \label{fig:xarm_tasks}
\end{figure*}

\subsection{Experimental Setup}

\subsubsection{Evaluation Benchmarks}
\textbf{LIBERO.}
LIBERO~\cite{liu2023libero} is a robotic manipulation benchmark designed to evaluate policy generalization across different task compositions, object configurations, and spatial arrangements. It comprises four evaluation suites (Spatial, Object, Goal, and Long), each targeting a distinct aspect of robotic generalization. We train a single model on all 1,693 training trajectories spanning 40 tasks.

\textbf{RoboTwin 2.0.}
RoboTwin 2.0~\cite{chen2025robotwin} is a challenging simulated benchmark for robotic manipulation. We evaluate on the ALOHA embodiment across all 50 tasks. For each task, the benchmark provides 50 demonstrations under clean scenes and 500 demonstrations under domain-randomized scenes (with randomization along five axes: clutter, lighting, background, tabletop height, and language instructions), yielding 27,500 training demonstrations in total. We merge the clean and randomized demonstrations and train a single model across all 50 tasks.

\textbf{RoboCasa GR-1 Tabletop.}
We also introduce RoboCasa GR-1 Tabletop as a evaluation benchmark built upon the RoboCasa simulation platform~\cite{robocasa2024}. This benchmark deploys the GR-1 humanoid robot in simulated tabletop environments, comprising 24 manipulation tasks that cover common sensorimotor skills such as picking, placing, and manipulating household objects. The use of a humanoid embodiment provides a complementary evaluation axis to the single-arm (LIBERO) and bimanual (RoboTwin 2.0) settings. We train a single model across all 24 tasks.

\textbf{Real-World (xArm).}
We conduct real-world experiments on an xArm robotic arm. We collect over 2,000 teleoperated trajectories spanning 4 manipulation tasks with 50+ distinct objects at 5\,Hz control frequency. The tasks include Push Blocks, Clean Table, Pick \& Place, and Hang Cups, as illustrated in Figure~\ref{fig:xarm_tasks}. During evaluation, each trial uses different object placements, task instructions (selected from the training set or freely rephrased), and randomly placed distractor objects, to evaluate the model's generalization under out-of-distribution conditions.

\begin{table}[t]
\centering
\small
\caption{Evaluation results on LIBERO (Success Rate, \%).}
\label{tab:libero_results}
\begin{tabular}{lccccc}
\toprule
Model & LIBERO-Spatial & LIBERO-Object & LIBERO-Goal & LIBERO-10 & Avg. \\
\midrule
OpenVLA~\cite{kim2024@openvla} 
& 84.7 & 88.4 & 79.2 & 53.7 & 76.5 \\

CoT-VLA~\cite{zhao2025cot} 
& 87.5 & 91.6 & 87.6 & 69.0 & 83.9 \\

$\pi_{0}$~\cite{black2024@pi0} 
& 96.8 & 98.8 & 95.8 & 85.2 & 94.2 \\

NORA-1.5~\cite{hung2025nora} 
& 97.3 & 96.4 & 94.5 & 89.6 & 94.5 \\

$\pi_{0.5}$~\cite{black2025@pi0.5} 
& 98.8 & 98.2 & 98.0 & 92.4 & 96.9 \\

GR00T-N1.7~\cite{bjorck2025gr00t} 
& 97.7 & 97.5 & 98.5 & 94.4 & 97.0 \\

DeepThinkVLA~\cite{yin2025deepthinkvla} 
& 96.6 & 99.0 & 96.4 & 96.2 & 97.0 \\

MolmoAct2~\cite{fang2026molmoact2} 
& 97.8 & 100.0 & 97.8 & 93.2 & 97.2 \\

\midrule
\model 
& 97.4 & 99.4 & 98.0 & 96.4 & \textbf{97.8} \\
\bottomrule
\end{tabular}
\end{table}

\subsubsection{Evaluation metric.}
We report the success rate (SR) as the main metric across three simulation benchmarks. For each episode, the score is binary: 1 if the task is successfully completed and 0 otherwise. We evaluate with 50 episodes per task for LIBERO, and 100 episodes per task for RoboTwin 2.0 (under both clean and randomized settings) and RoboCasa GR-1 Tabletop. For real-world experiments, we conduct 10 trials per task and adopt a task progress score $S \in [0, 100]$, where each task is decomposed into a sequence of sub-steps and scored according to a task-specific rubric.

\subsubsection{Implementation Details}
\model comprises approximately 2.6 billion parameters in total: 2.1 billion in the VLM (initialized from Qwen3-VL-2B-Instruct) and 500 million in the DiT-based action expert. During pre-training, the action chunk length is $H=32$, the global batch size is 1,024, and the loss weight is $\alpha=5$. To accommodate the variability in state and action dimensions across embodiments, we pad both states and actions to 64 dimensions with zeros, and mask the loss on padded dimensions so that they do not contribute gradients during training. Each dimension is min-max normalized using the 1st and 99th percentiles of the training data. We use the AdamW optimizer~\citep{Loshchilov2019@adamw} with $\beta_1{=}0.9$, $\beta_2{=}0.95$, and $\epsilon{=}10^{-8}$. The learning rate follows a cosine schedule with a linear warm-up over the first 5\% of steps, ramping from $0$ to a peak of $3 \times 10^{-5}$ and decaying to $3 \times 10^{-6}$. Training uses \texttt{bfloat16} mixed precision with gradient clipping at 1.0. We employ DeepSpeed ZeRO~\citep{Rajbhandari2020@zero} for memory-efficient distributed training, together with Flash-Attention 2~\citep{dao2024@flash-attn2} and gradient checkpointing to further reduce memory consumption.

\subsubsection{Post-training and inference setting.}
We post-train \model on each benchmark's training demonstrations with a batch size of 64, a loss weight of $\alpha=1$, and an action chunk length of $H=10$ for LIBERO and $H=16$ for RoboTwin 2.0, RoboCasa GR-1 Tabletop, and real-world xArm. To ensure a fair comparison with baseline methods, the post-training stage follows a standard protocol using only the publicly available benchmark training data, without any ECoT supervision or VL data co-training. Both ECoT and VL data are used exclusively during pre-training to improve the cross-embodiment representations learned by the VLM. At inference time, \model generates an action chunk via flow matching, executes it in the environment, and then re-plans from the latest observations.

\begin{table*}[t]
\centering
\small
\setlength{\tabcolsep}{3.2pt}
\caption{Evaluation results on the RoboCasa GR-1 Tabletop benchmark (Success Rate, \%).}
\label{tab:robocasa_results}
\begin{tabular}{lcccccc}
\toprule
\multirow{2}{*}{Task}
& GR00T
& \multirow{2}{*}{Qwen3PI~\cite{community2026starvla}}
& VP
& ABot
& JoyAI
& \multirow{2}{*}{\model} \\
& -N1.6~\cite{bjorck2025gr00t} & & -VLA~\cite{wang2026vp} & -M0~\cite{yang2026abot} & -RA~\cite{zhang2026joyai} & \\
\midrule
BottleToCabinetClose        & 51.5 & 26.0 & 54.0 & 86.0 & 84.0 & 39.0 \\
CanToDrawerClose            & 13.0 & 62.0 & 72.0 & 74.0 & 90.0 & 47.0 \\
CupToDrawerClose            & 8.5  & 42.0 & 44.0 & 48.0 & 48.0 & 20.0 \\
MilkToMicrowaveClose        & 14.0 & 50.0 & 74.0 & 46.0 & 84.0 & 45.0 \\
PotatoToMicrowaveClose      & 41.5 & 42.0 & 34.0 & 50.0 & 70.0 & 59.0 \\
WineToCabinetClose          & 16.5 & 32.0 & 48.0 & 66.0 & 54.0 & 40.0 \\
CuttingboardToBasket        & 58.0 & 40.0 & 66.0 & 70.0 & 88.0 & 82.0 \\
CuttingboardToCardboardbox  & 46.5 & 46.0 & 54.0 & 58.0 & 46.0 & 79.0 \\
CuttingboardToPan           & 68.5 & 60.0 & 74.0 & 76.0 & 92.0 & 92.0 \\
CuttingboardToPot           & 65.0 & 40.0 & 54.0 & 66.0 & 80.0 & 85.0 \\
CuttingboardToTieredbasket  & 46.5 & 44.0 & 56.0 & 38.0 & 36.0 & 80.0 \\
PlacematToBasket            & 58.5 & 44.0 & 48.0 & 52.0 & 76.0 & 78.0 \\
PlacematToBowl              & 57.5 & 52.0 & 74.0 & 66.0 & 52.0 & 87.0 \\
PlacematToPlate             & 63.0 & 50.0 & 70.0 & 60.0 & 38.0 & 88.0 \\
PlacematToTieredshelf       & 28.5 & 28.0 & 26.0 & 26.0 & 14.0 & 46.0 \\
PlateToBowl                 & 57.0 & 52.0 & 52.0 & 54.0 & 48.0 & 82.0 \\
PlateToCardboardbox         & 43.5 & 40.0 & 44.0 & 48.0 & 38.0 & 81.0 \\
PlateToPan                  & 51.0 & 36.0 & 56.0 & 66.0 & 46.0 & 89.0 \\
PlateToPlate                & 78.7 & 48.0 & 62.0 & 64.0 & 88.0 & 85.0 \\
TrayToCardboardbox          & 51.5 & 34.0 & 44.0 & 54.0 & 82.0 & 81.0 \\
TrayToPlate                 & 71.0 & 64.0 & 66.0 & 68.0 & 88.0 & 81.0 \\
TrayToPot                   & 64.5 & 44.0 & 38.0 & 64.0 & 88.0 & 74.0 \\
TrayToTieredbasket          & 57.0 & 50.0 & 58.0 & 60.0 & 62.0 & 74.0 \\
TrayToTieredshelf           & 31.5 & 28.0 & 24.0 & 38.0 & 24.0 & 50.0 \\
\midrule
Average                     & 47.6 & 43.9 & 53.8 & 58.3 & 63.2 & \textbf{69.3} \\
\bottomrule
\end{tabular}
\end{table*}

\subsection{Experimental Results}

\subsubsection{Simulation Experiments}

Table~\ref{tab:libero_results}, \ref{tab:robocasa_results}, and \ref{tab:robotwin_results} summarize the simulation results across three embodiments. \model achieves competitive performance on all three benchmarks: 97.8\% on LIBERO, 69.3\% on RoboCasa GR-1 Tabletop, and 88.70\%/87.98\% (Clean/Randomized) on RoboTwin 2.0. These results span single-arm, humanoid, and bimanual embodiments, all fine-tuned from the same pre-trained checkpoint, supporting the effectiveness of ECoT-supervised pre-training for cross-embodiment adaptation.

On LIBERO, the performance gap between methods is most visible on LIBERO-10, the long-horizon suite that chains multiple manipulation sub-goals. \model reaches 96.4\% on this suite, 4.0 points above $\pi_{0.5}$. The other three suites are largely saturated across recent methods, making LIBERO-10 the primary differentiating factor.

On RoboCasa GR-1 Tabletop, \model achieves 69.3\% average success rate, outperforming the next best method (JoyAI-RA, 63.2\%) by 6.1 points. Compared with JoyAI-RA, \model shows substantial improvements on pick-and-place tasks, for example CuttingboardToTieredbasket (80\% vs.\ 36\%), PlacematToPlate (88\% vs.\ 38\%), PlateToPan (89\% vs.\ 46\%), and PlateToBowl (82\% vs.\ 48\%). However, \model underperforms on the six Close tasks that require multi-phase interactions with cabinets, drawers, and microwaves (e.g., BottleToCabinetClose 39\% vs.\ 84\%, CanToDrawerClose 47\% vs.\ 90\%). These results suggest that pick-and-place, the most prevalent manipulation primitive in the pre-training corpus, benefits most from ECoT-supervised representation alignment and transfers effectively to downstream tasks. In contrast, closing actions (e.g., shutting cabinets, drawers, and microwaves) appear far less frequently in the pre-training data, limiting the model's ability to learn well-aligned representations for these behaviors.

On RoboTwin 2.0, \model achieves slightly better performance than LingBot-VLA while using a substantially smaller pre-training corpus (approximately 1,000 hours versus 20,000 hours). \model reaches at or near 100\% success rate on several tasks under both Clean and Randomized settings (GrabRoller, HandoverMic, ShakeBottle). On tasks that specifically require bimanual coordination (HandoverBlock 93/87\%, HandoverMic 100/99\%, PickDualBottles 97/98\%), \model also demonstrates strong performance. On multi-step tasks that require three sequential operations (e.g., BlocksRankingRGB 92/91\%, StackBlocksThree 86/88\%, StackBowlsThree 79/88\%), \model also shows competitive results, with notably higher Randomized scores than Clean on some tasks (e.g., BlocksRankingSize 70$\to$81\%, StackBowlsThree 79$\to$88\%), likely benefiting from the diverse open-world scenes encountered through VL data co-training during pre-training. More broadly, the performance gap between Clean and Randomized settings is small across the board: \model drops only 0.72 points on average, compared to 1.64 for Motus and 5.98 for $\pi_{0.5}$, suggesting stronger robustness to visual variations in clutter, lighting, and background.

\begin{table*}[t]
\centering
\small
\setlength{\tabcolsep}{2.4pt}
\caption{Evaluation results on RoboTwin2.0. We report the success rate (SR, \%). Full per-task results are provided in Table~\ref{tab:robotwin_results_full} in the Appendix. LB-VLA is the abbreviation of LingBot-VLA.}
\label{tab:robotwin_results}
\begin{tabular}{l*{12}{c}}
\toprule
Task
& \multicolumn{2}{c}{$\pi_0$~\cite{black2024@pi0}}
& \multicolumn{2}{c}{$\pi_{0.5}$~\cite{black2025@pi0.5}}
& \multicolumn{2}{c}{X-VLA~\cite{zheng2025x}}
& \multicolumn{2}{c}{Motus~\cite{bi2026motus}}
& \multicolumn{2}{c}{LB-VLA~\cite{wu2026pragmatic}}
& \multicolumn{2}{c}{\model} \\
\cmidrule(lr){2-3}
\cmidrule(lr){4-5}
\cmidrule(lr){6-7}
\cmidrule(lr){8-9}
\cmidrule(lr){10-11}
\cmidrule(lr){12-13}
& Clean & Rand.
& Clean & Rand.
& Clean & Rand.
& Clean & Rand.
& Clean & Rand.
& Clean & Rand. \\
\midrule
AdjustBottle              & 99 & 95 & 100 & 99 & 100 & 99 & 89 & 93 & 100 & 100 & 100 & 99 \\
BeatBlockHammer           & 79 & 84 & 96 & 93 & 92 & 88 & 95 & 88 & 92 & 89 & 85 & 88 \\
BlocksRankingRGB          & 80 & 63 & 92 & 85 & 83 & 83 & 99 & 97 & 92 & 91 & 92 & 91 \\
BlocksRankingSize         & 14 & 5  & 49 & 26 & 67 & 74 & 75 & 63 & 76 & 70 & 70 & 81 \\
ClickAlarmclock           & 77 & 68 & 98 & 89 & 99 & 99 & 100 & 100 & 97 & 43 & 96 & 82 \\
ClickBell                 & 71 & 48 & 99 & 66 & 100 & 100 & 100 & 100 & 43 & 36 & 90 & 83 \\
DumpBinBigbin             & 88 & 83 & 92 & 97 & 79 & 77 & 95 & 91 & 97 & 97 & 94 & 93 \\
GrabRoller                & 98 & 94 & 100 & 100 & 100 & 100 & 100 & 100 & 100 & 100 & 100 & 100 \\
HandoverBlock             & 47 & 31 & 66 & 57 & 73 & 37 & 86 & 73 & 99 & 93 & 93 & 87 \\
HandoverMic               & 97 & 97 & 98 & 97 & 0 & 0 & 78 & 63 & 100 & 99 & 100 & 99 \\
\multicolumn{13}{c}{$\vdots$} \\
StackBlocksThree          & 72 & 52 & 91 & 76 & 6 & 10 & 91 & 95 & 60 & 62 & 86 & 88 \\
StackBlocksTwo            & 93 & 79 & 97 & 100 & 92 & 87 & 100 & 98 & 95 & 93 & 95 & 91 \\
StackBowlsThree           & 77 & 75 & 77 & 71 & 76 & 86 & 79 & 87 & 80 & 81 & 79 & 88 \\
StackBowlsTwo             & 94 & 95 & 95 & 96 & 96 & 93 & 98 & 98 & 95 & 93 & 94 & 92 \\
StampSeal                 & 46 & 33 & 79 & 55 & 76 & 82 & 93 & 92 & 90 & 90 & 92 & 92 \\
TurnSwitch                & 41 & 42 & 62 & 54 & 40 & 61 & 84 & 78 & 71 & 76 & 83 & 78\\
\midrule
Average
& 65.92 & 58.40
& 82.74 & 76.76
& 72.80 & 72.84
& 88.66 & 87.02
& 88.56 & 86.68
& \textbf{88.70} & \textbf{87.98} \\
\bottomrule
\end{tabular}
\end{table*}

\subsubsection{Real-World Experiments}
We compare \model with $\pi_{0.5}$ on four real-world xArm tasks, each designed to probe a different capability. As shown in Table~\ref{tab:xarm_results}, \model achieves an average task progress score of 76.0, outperforming $\pi_{0.5}$ (67.8) by 8.2 points.

\textbf{Push Blocks} tests fine-grained manipulation of small objects and OCR-based reasoning, as the model must read letters printed on wooden blocks to identify the correct targets. \model scores 94.0 on this task, a 27.9-point improvement over $\pi_{0.5}$ (66.1), the largest gain among the four tasks. We attribute this to VL data co-training and ECoT reasoning, which preserves the VLM's original text recognition capability that would otherwise degrade under action-only fine-tuning. \textbf{Clean Table} evaluates long-horizon execution, requiring the model to repeatedly pick up objects and place them into a designated area across many sequential steps. \model scores 73.4 versus 63.3 for $\pi_{0.5}$. This improvement aligns with the role of the To-Do Actions component in ECoT, which decomposes long-horizon goals into atomic, embodiment-agnostic sub-tasks and provides explicit alignment for the pick-and-place primitive. \textbf{Pick \& Place} focuses on spatial reasoning and referential language understanding (e.g., ``Put the green apple on the plate with a banana.''), where \model scores 66.7 versus 56.7 for $\pi_{0.5}$. This gain can be traced to ECoT's Scene Description and Target Objects components, which train the VLM to perceive object spatial relationships and ground task-relevant regions in the visual observation.

On \textbf{Hang Cups}, which requires color understanding to identify the target cup and precise dexterous control to align and hang it on a hook, $\pi_{0.5}$ outperforms \model (85.0 vs.\ 70.0). This task demands fine-grained motor precision that goes beyond high-level reasoning, suggesting that while ECoT supervision strengthens scene understanding and planning, highly precise manipulation may depend more on the scale of action supervision during pre-training.

\begin{table}[t]
\centering
\caption{Experimental results on real-world xArm platform (Progress Score).}
\label{tab:xarm_results}
\begin{tabular}{lccccc}
\toprule
Method & Pick \& Place & Hang Cups & Clean Table & Push Blocks & Avg. \\
\midrule
$\pi_{0.5}$ & 56.7 & \textbf{85.0} & 63.3 & 66.1 & 67.8 \\
\model & \textbf{66.7} & 70.0 & \textbf{73.4} & \textbf{94.0} & \textbf{76.0} \\
\bottomrule
\end{tabular}
\end{table}

\begin{table}[t]
\centering
\caption{Ablation study on LIBERO (Success Rate, \%).}
\label{tab:libero_ablation}
\begin{tabular}{lccccc}
\toprule
Method & LIBERO-Spatial & LIBERO-Object & LIBERO-Goal & LIBERO-10 & Avg. \\
\midrule
\model & \textbf{97.4} & \textbf{99.4} & \textbf{98.0} & \textbf{96.4} & \textbf{97.8} \\
- w/o ECoT PT & 96.8 & 98.6 & 94.8 & 92.6 & 95.7 \\
\bottomrule
\end{tabular}
\end{table}

\subsection{Ablation study.}

We study the effect of ECoT-supervised pre-training on LIBERO in Table~\ref{tab:libero_ablation}. The \textbf{w/o ECoT PT} baseline initializes the VLM from the Qwen3-VL-2B-Instruct base model with a randomly initialized action expert and directly fine-tunes on LIBERO, bypassing the ECoT-supervised pre-training stage entirely. Both settings use the same post-training configuration described above. The results show that removing ECoT-supervised pre-training leads to a clear drop in success rate, confirming that the cross-embodiment representations learned during pre-training transfer effectively to downstream tasks. ECoT pre-training provides the VLM with structured supervision for scene understanding, task progress estimation, future planning, and target-object grounding, resulting in stronger initial representations that facilitate more efficient downstream adaptation.

\section{Discussion}

\textbf{Scaling Robot Data.}
Despite the strong results presented in this work, our current pre-training corpus comprises approximately 1,000 hours of robot data, which is an order of magnitude below leading VLA models such as $\pi_{0}$~\citep{black2024@pi0} (over 10,000 hours), LingBot-VLA~\citep{wu2026pragmatic} (around 20,000 hours) and Qwen-RobotManip~\citep{yuan2026@qwenrobomanip} (over 30,000 hours). As shown in our RoboCasa experiments, skills that are well-represented in the pre-training data (e.g., pick-and-place) benefit significantly from ECoT-supervised representation alignment, while underrepresented skills (e.g., closing cabinets and drawers) show weaker adaptation. Scaling the pre-training corpus to cover a broader range of manipulation primitives would directly expand the set of skills for which ECoT can learn aligned, transferable representations. Moreover, as suggested by our real-world Hang Cups results, increasing the scale of action supervision during pre-training may also improve fine-grained motor precision for tasks that demand dexterous control beyond high-level reasoning.

\textbf{Learning from Human Egocentric Video.}
A distinctive property of ECoT is that its structured reasoning (scene descriptions, task planning, sub-task decomposition, and object grounding) is agnostic to whether the manipulation is performed by a robot or a human. This opens a promising avenue for leveraging the vast body of human egocentric video data (e.g., Ego4D~\cite{Grauman2022@ego4D}, EPIC-KITCHENS~\cite{Damen2021@EPIC-KITCHENS}) to enhance VLA pre-training. By annotating human manipulation videos with ECoT, the VLM can acquire richer visual and semantic representations of manipulation behaviors at a scale that robot-only data cannot yet provide, without requiring any robot action labels.

\textbf{Efficient ECoT Annotation.}
Annotating every frame in a robot trajectory with dense ECoT requires substantial computational resources, as each annotation involves a forward pass through a capable VLM. A promising direction for future research is developing strategies to select the most informative frames for ECoT annotation, rather than annotating exhaustively. The goal is to match the representation quality achieved by dense annotation while significantly reducing the annotation cost, making ECoT-based pre-training more scalable and accessible.
\section{Conclusion}

We present \model, a 2.6B parameter VLA model that uses dense Embodied Chain-of-Thought supervision to align cross-embodiment representations within the VLM. By coupling a pre-trained VLM with a DiT-based action expert through cross-attention and restricting the action expert to input prompt features only, \model benefits from ECoT's rich training signal while entirely skipping ECoT generation at inference, achieving approximately 100\,ms per action chunk on a single H100 GPU. Pre-trained on ProcCorpus-60M (approximately 1,000 hours, 96.8\% ECoT annotation coverage), \model achieves 97.8\% on LIBERO, 69.3\% on RoboCasa GR-1 Tabletop, and 88.70\%/87.98\% (Clean/Randomized) on RoboTwin 2.0, all fine-tuned from the same pre-trained checkpoint. Real-world xArm experiments further confirm the benefits of ECoT supervision for scene understanding, spatial reasoning, and long-horizon planning. We hope that \model demonstrates the potential of structured reasoning supervision as a scalable and embodiment-agnostic approach to cross-embodiment representation learning for VLA models.

\clearpage
\bibliographystyle{plainnat}
\bibliography{neurips_2025}

\appendix

\section{Full RoboTwin 2.0 Results}

Table~\ref{tab:robotwin_results_full} presents the complete per-task evaluation results on all 50 RoboTwin 2.0 tasks under both Clean and Randomized settings. The abbreviated version in the main text (Table~\ref{tab:robotwin_results}) includes a representative subset; this table provides the full breakdown for reference.

\begin{table*}[t]
\centering
\small
\setlength{\tabcolsep}{1.6pt}
\caption{Evaluation results on RoboTwin2.0 (Success Rate, \%).}
\label{tab:robotwin_results_full}
\begin{tabular}{l*{12}{c}}
\toprule
Task 
& \multicolumn{2}{c}{$\pi_0$~\cite{black2024@pi0}} 
& \multicolumn{2}{c}{$\pi_{0.5}$~\cite{black2025@pi0.5}} 
& \multicolumn{2}{c}{X-VLA~\cite{zheng2025x}} 
& \multicolumn{2}{c}{Motus~\cite{bi2026motus}} 
& \multicolumn{2}{c}{LingBot-VLA~\cite{wu2026pragmatic}} 
& \multicolumn{2}{c}{\model} \\
\cmidrule(lr){2-3} 
\cmidrule(lr){4-5} 
\cmidrule(lr){6-7} 
\cmidrule(lr){8-9} 
\cmidrule(lr){10-11} 
\cmidrule(lr){12-13}
& Easy & Hard 
& Easy & Hard 
& Easy & Hard 
& Easy & Hard 
& Easy & Hard 
& Easy & Hard \\
\midrule
AdjustBottle              & 99 & 95 & 100 & 99 & 100 & 99 & 89 & 93 & 100 & 100 & 100 & 99 \\
BeatBlockHammer           & 79 & 84 & 96 & 93 & 92 & 88 & 95 & 88 & 92 & 89 & 85 & 88 \\
BlocksRankingRGB          & 80 & 63 & 92 & 85 & 83 & 83 & 99 & 97 & 92 & 91 & 92 & 91 \\
BlocksRankingSize         & 14 & 5  & 49 & 26 & 67 & 74 & 75 & 63 & 76 & 70 & 70 & 81 \\
ClickAlarmclock           & 77 & 68 & 98 & 89 & 99 & 99 & 100 & 100 & 97 & 43 & 96 & 82 \\
ClickBell                 & 71 & 48 & 99 & 66 & 100 & 100 & 100 & 100 & 43 & 36 & 90 & 83 \\
DumpBinBigbin             & 88 & 83 & 92 & 97 & 79 & 77 & 95 & 91 & 97 & 97 & 94 & 93 \\
GrabRoller                & 98 & 94 & 100 & 100 & 100 & 100 & 100 & 100 & 100 & 100 & 100 & 100 \\
HandoverBlock             & 47 & 31 & 66 & 57 & 73 & 37 & 86 & 73 & 99 & 93 & 93 & 87 \\
HandoverMic               & 97 & 97 & 98 & 97 & 0 & 0 & 78 & 63 & 100 & 99 & 100 & 99 \\
HangingMug                & 14 & 11 & 18 & 17 & 23 & 27 & 38 & 38 & 31 & 28 & 35 & 33 \\
LiftPot                   & 80 & 72 & 96 & 85 & 99 & 100 & 96 & 99 & 100 & 99 & 96 & 98 \\
MoveCanPot                & 68 & 48 & 51 & 55 & 89 & 86 & 34 & 74 & 97 & 87 & 85 & 81 \\
MovePillbottlePad         & 67 & 46 & 84 & 61 & 73 & 71 & 93 & 96 & 98 & 99 & 98 & 99 \\
MovePlayingcardAway       & 74 & 65 & 96 & 84 & 93 & 98 & 100 & 96 & 99 & 95 & 99 & 94 \\
MoveStaplerPad            & 41 & 24 & 56 & 42 & 78 & 73 & 83 & 85 & 93 & 96 & 85 & 92 \\
OpenLaptop                & 71 & 81 & 90 & 96 & 93 & 100 & 95 & 91 & 96 & 100 & 96 & 99 \\
OpenMicrowave             & 4  & 32 & 34 & 77 & 79 & 71 & 95 & 91 & 97 & 99 & 94 & 92 \\
PickDiverseBottles        & 69 & 31 & 81 & 71 & 58 & 36 & 90 & 91 & 85 & 90 & 90 & 88 \\
PickDualBottles           & 59 & 37 & 93 & 63 & 47 & 36 & 96 & 90 & 95 & 93 & 97 & 98 \\
PlaceA2BLeft              & 43 & 47 & 87 & 82 & 48 & 49 & 82 & 79 & 99 & 96 & 83 & 80 \\
PlaceA2BRight             & 39 & 34 & 87 & 84 & 36 & 36 & 90 & 87 & 97 & 92 & 86 & 87 \\
PlaceBreadBasket          & 62 & 46 & 77 & 64 & 81 & 71 & 91 & 94 & 88 & 91 & 90 & 93 \\
PlaceBreadSkillet         & 66 & 49 & 85 & 66 & 77 & 67 & 86 & 83 & 92 & 89 & 92 & 85 \\
PlaceBurgerFries          & 81 & 76 & 94 & 87 & 94 & 94 & 98 & 98 & 99 & 93 & 98 & 98 \\
PlaceCanBasket            & 55 & 46 & 62 & 62 & 49 & 52 & 81 & 76 & 71 & 73 & 66 & 62 \\
PlaceCansPlasticbox       & 63 & 45 & 94 & 84 & 97 & 98 & 98 & 94 & 100 & 98 & 86 & 85 \\
PlaceContainerPlate       & 97 & 92 & 99 & 95 & 97 & 95 & 98 & 99 & 96 & 99 & 99 & 98 \\
PlaceDualShoes            & 59 & 51 & 75 & 75 & 79 & 88 & 93 & 87 & 90 & 97 & 90 & 95 \\
PlaceEmptyCup             & 91 & 85 & 100 & 99 & 100 & 98 & 99 & 98 & 100 & 100 & 97 & 97 \\
PlaceFan                  & 66 & 71 & 87 & 85 & 80 & 75 & 91 & 87 & 91 & 92 & 85 & 78 \\
PlaceMousePad             & 20 & 20 & 60 & 39 & 70 & 70 & 66 & 68 & 89 & 82 & 85 & 83\\
PlaceObjectBasket         & 67 & 70 & 80 & 76 & 44 & 39 & 81 & 87 & 90 & 88 & 75 & 77 \\
PlaceObjectScale          & 57 & 52 & 86 & 80 & 52 & 74 & 88 & 85 & 90 & 87 & 88 & 89 \\
PlaceObjectStand          & 82 & 68 & 91 & 85 & 86 & 88 & 98 & 97 & 95 & 93 & 90 & 91 \\
PlacePhoneStand           & 49 & 53 & 81 & 81 & 88 & 87 & 87 & 86 & 95 & 95 & 85 & 81 \\
PlaceShoe                 & 76 & 76 & 92 & 93 & 96 & 95 & 99 & 97 & 99 & 100 & 98 & 97 \\
PressStapler              & 44 & 37 & 87 & 83 & 92 & 98 & 93 & 98 & 87 & 81 & 90 & 92 \\
PutBottlesDustbin         & 65 & 56 & 84 & 79 & 74 & 77 & 81 & 79 & 95 & 97 & 82 & 79 \\
PutObjectCabinet          & 73 & 60 & 80 & 79 & 46 & 48 & 88 & 71 & 87 & 86 & 82 & 76 \\
RotateQRcode              & 74 & 70 & 89 & 87 & 34 & 33 & 89 & 73 & 83 & 82 & 78 & 85 \\
ScanObject                & 55 & 42 & 72 & 65 & 14 & 36 & 67 & 66 & 98 & 96 & 86 & 85 \\
ShakeBottleHorizontally   & 98 & 92 & 99 & 99 & 100 & 100 & 100 & 98 & 100 & 100 &100 & 100 \\
ShakeBottle               & 94 & 91 & 99 & 97 & 99 & 100 & 100 & 97 & 100 & 100 & 100 & 100 \\
StackBlocksThree          & 72 & 52 & 91 & 76 & 6 & 10 & 91 & 95 & 60 & 62 & 86 & 88 \\
StackBlocksTwo            & 93 & 79 & 97 & 100 & 92 & 87 & 100 & 98 & 95 & 93 & 95 & 91 \\
StackBowlsThree           & 77 & 75 & 77 & 71 & 76 & 86 & 79 & 87 & 80 & 81 & 79 & 88 \\
StackBowlsTwo             & 94 & 95 & 95 & 96 & 96 & 93 & 98 & 98 & 95 & 93 & 94 & 92 \\
StampSeal                 & 46 & 33 & 79 & 55 & 76 & 82 & 93 & 92 & 90 & 90 & 92 & 92 \\
TurnSwitch                & 41 & 42 & 62 & 54 & 40 & 61 & 84 & 78 & 71 & 76 & 83 & 78\\
\midrule
Average 
& 65.92 & 58.40 
& 82.74 & 76.76 
& 72.80 & 72.84 
& 88.66 & 87.02 
& 88.56 & 86.68 
& \textbf{88.70} & \textbf{87.98} \\
\bottomrule
\end{tabular}
\end{table*}

\end{document}